\documentclass[10pt,twocolumn,letterpaper]{article}

\usepackage{authblk}

\usepackage[pagenumbers]{cvpr} 

\usepackage[dvipsnames]{xcolor}
\usepackage{graphicx}
\graphicspath{ {./images/} }
\usepackage{amsmath}
\usepackage{amssymb}
\usepackage{comment}
\usepackage{booktabs}
\usepackage[setmargin=true,marginparwidth=0.4in,hide=false]{marginalia}

\usepackage[pagebackref,breaklinks,colorlinks]{hyperref}

\usepackage[capitalize]{cleveref}
\crefname{section}{Sec.}{Secs.}
\Crefname{section}{Section}{Sections}
\Crefname{table}{Table}{Tables}
\crefname{table}{Tab.}{Tabs.}

\DeclareMathOperator*{\argmin}{arg\,min}

\def\cvprPaperID{8247} 
\def\confName{CVPR}
\def\confYear{2022}

\title{3D Neural Field Generation using Triplane Diffusion}

\makeatletter
    \renewcommand\AB@affilsepx{ \hphantom{---} \protect\Affilfont}
\makeatother

\author{J. Ryan Shue\thanks{Equal contribution.} $^1$ $\,\,$ Eric Ryan Chan$^{*2}$ $\,\,$ Ryan Po$^{*2}$ $\,\,$ Zachary Ankner$^{*3,4}$ $\,\,$ Jiajun Wu$^2$ $\,\,$ Gordon Wetzstein$^2$\\
$^1$Milton Academy $\,\,\,$ $^2$Stanford University $\,\,\,$ $^3$Massachusetts Institute of Technology $\,\,\,$ $^4$MosaicML
}

\begin{document}

\maketitle

\begin{abstract}
    Diffusion models have emerged as the state-of-the-art for image generation, among other tasks. Here, we present an efficient diffusion-based model for 3D-aware generation of neural fields. Our approach pre-processes training data, such as ShapeNet meshes, by converting them to continuous occupancy fields and factoring them into a set of axis-aligned triplane feature representations. Thus, our 3D training scenes are all represented by 2D feature planes, and we can directly train existing 2D diffusion models on these representations to generate 3D neural fields with high quality and diversity, outperforming alternative approaches to 3D-aware generation. Our approach requires essential modifications to existing triplane factorization pipelines to make the resulting features easy to learn for the diffusion model. We demonstrate state-of-the-art results on 3D generation on several object classes from ShapeNet.

\newcommand\blfootnote[1]{%
  \begingroup
  \renewcommand\thefootnote{}\footnote{#1}%
  \addtocounter{footnote}{-1}%
  \endgroup
}

\blfootnote{Part of the work was done during an internship at Stanford.}
\blfootnote{Project page: \href{https://jryanshue.com/nfd}{https://jryanshue.com/nfd}}


\end{abstract}

\section{Introduction}

Diffusion models have seen rapid progress, setting state-of-the-art (SOTA) performance across a variety of image generation tasks.
While most diffusion methods model 2D images, recent work \cite{Luo2021ddpm-point-cloud,Zhou2021point-voxel-ddpm,dupont2022functa,Bautista2022gaudi} has attempted to develop denoising methods for 3D shape generation. These 3D diffusion methods operate on discrete point clouds and, while successful, exhibit limited quality and resolution.

In contrast to 2D diffusion, which directly leverages the image as the target for the diffusion process, it is not directly obvious how to construct such 2D targets in the case of 3D diffusion. Interestingly, recent work on 3D-aware generative adversarial networks (GANs) (see Sec.~\ref{sec:related} for an overview) has demonstrated impressive results for 3D shape generation using 2D generators. We build upon this idea of learning to generate triplane representations~\cite{Chan2022eg3d} that encode 3D scenes or radiance fields as a set of axis-aligned 2D feature planes. The structure of a triplane is analogous to that of a 2D image and can be used as part of a 3D generative method that leverages conventional 2D generator architectures.

\begin{figure}[t!]
  \includegraphics[width=0.5\textwidth]{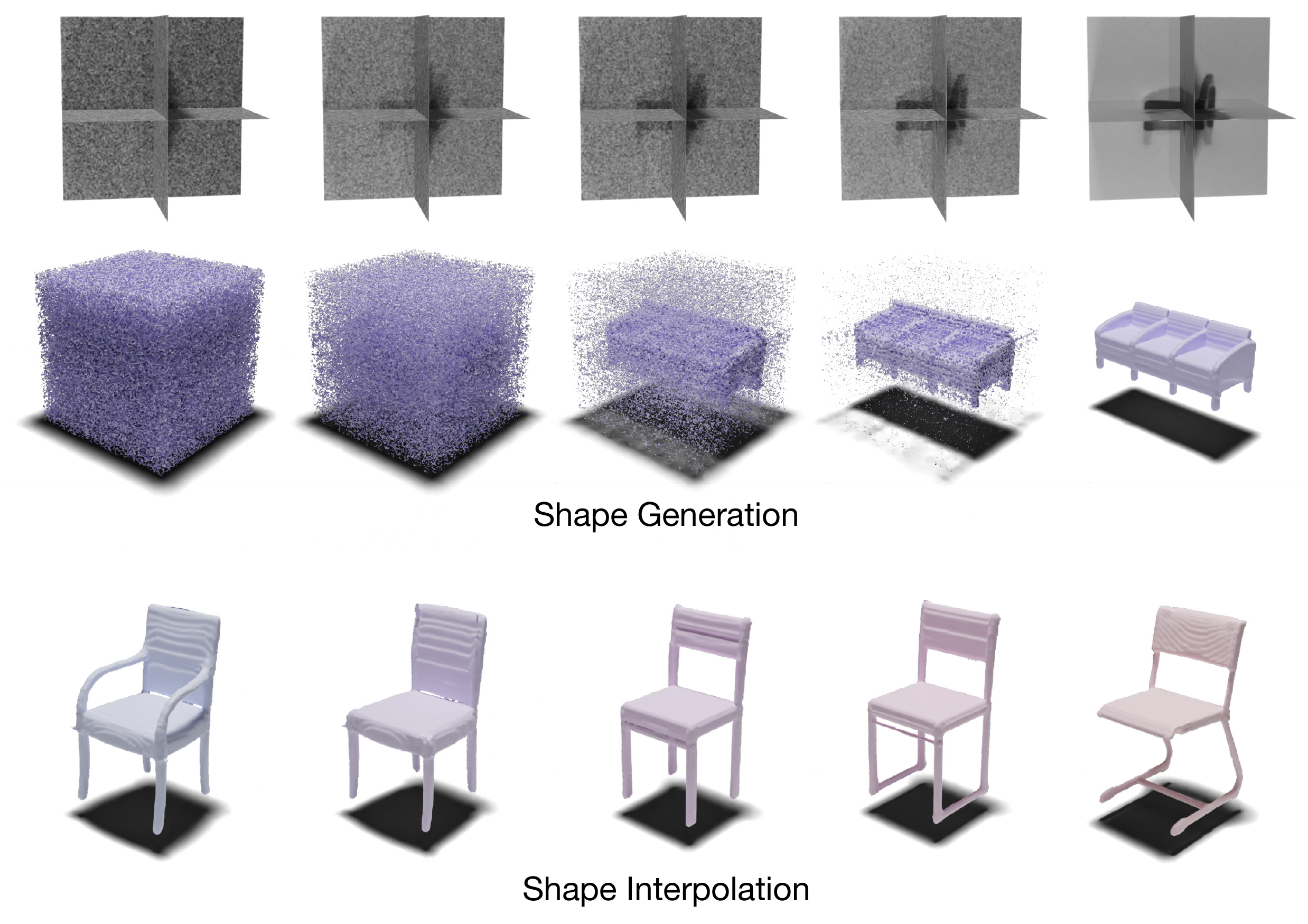}
  \caption{Our method leverages existing 2D diffusion models for 3D shape generation using hybrid explicit--implicit neural representations. Top: triplane-based 3D shape diffusion process using our framework. Bottom: Interpolation between generated shapes.}
  \label{fig:teaser}
\end{figure}

\begin{figure*}
    \centering
    \includegraphics[width=0.95\textwidth]{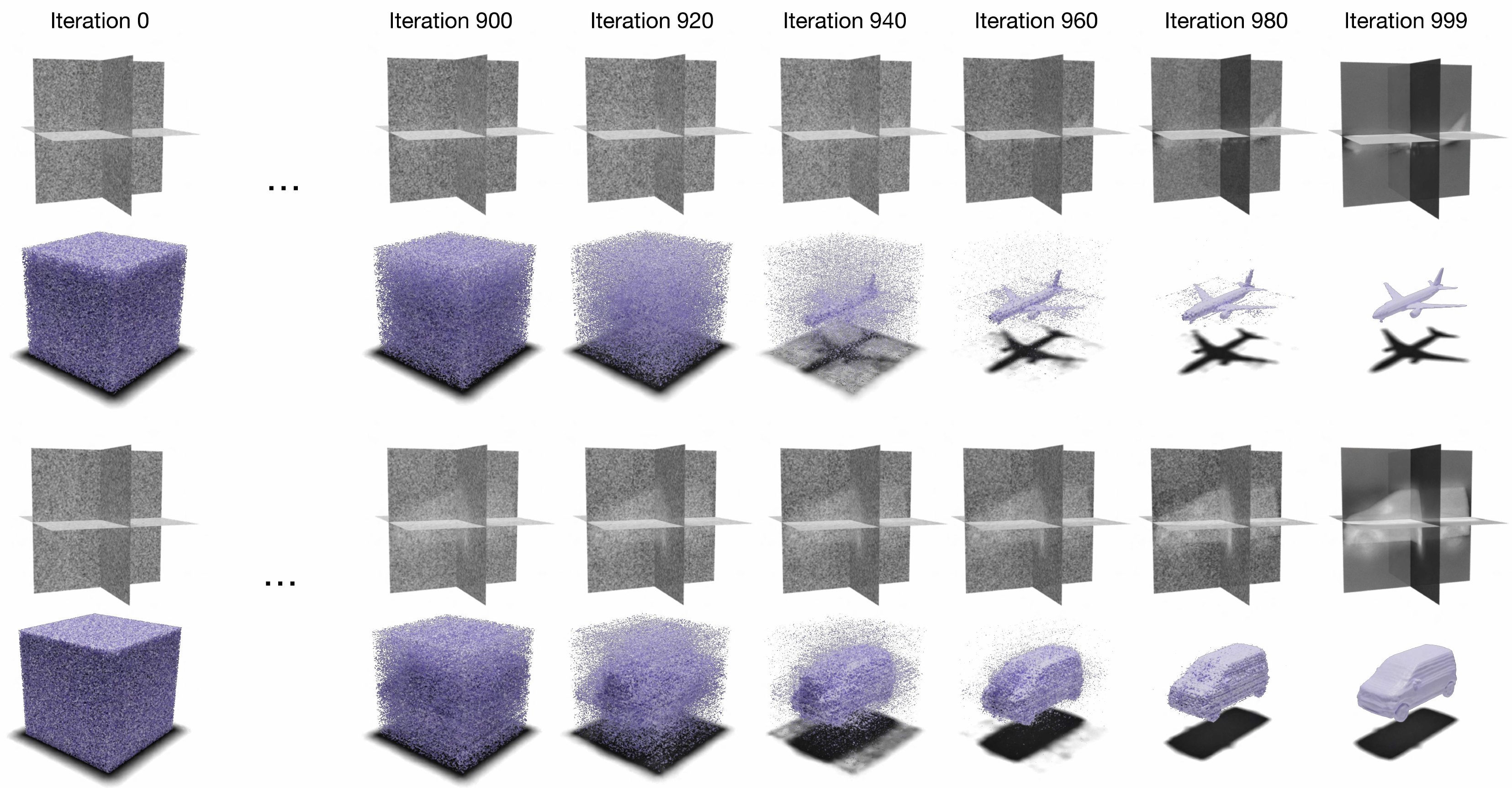}
    \caption{\textbf{Visualization of the denoising process.} Here, we show examples of triplanes as they are iteratively denoised at inference, as well as the shapes we obtain by ``decoding" the noisy triplanes with our jointly-learned MLP. By interpreting triplane features simply as multi-channel feature images, we build our framework around 2D diffusion models.}
    \label{fig:diffusion-process}
\end{figure*}

Inspired by recent efforts in designing efficient 3D GAN architectures, we introduce a neural field-based diffusion framework for 3D representation learning.
Our approach follows a two-step process. In the first step, a training set of 3D scenes is factored into a set of per-scene triplane features and a single, shared feature decoder. In the second step, a 2D diffusion model is trained on these triplanes. The trained diffusion model can then be used at inference time to generate novel and diverse 3D scenes. By interpreting triplanes as multi-channel 2D images and thus decoupling generation from rendering, we can leverage current (and likely future) SOTA 2D diffusion model backbones nearly out of the box. Fig.~\ref{fig:teaser} illustrates how a single object is generated with our framework (top), and how two generated objects---even with different topologies---can be interpolated (bottom).


Our core contributions are as follows:
\begin{itemize}
    \item We introduce a generative framework for diffusion on 3D scenes that utilizes 2D diffusion model backbones and has a built-in 3D inductive bias.
    \item We show that our approach is capable of generating both high-fidelity and diverse 3D scenes that outperform state-of-the-art 3D GANs.
\end{itemize}

\section{Related Work}
\label{sec:related}

\paragraph{Neural fields.}
Implicit neural representations, or neural fields, hold the SOTA for 3D scene representation \cite{xie2022neuralfields,tewari2022advances}.
They either solely learn geometry \cite{park2019deepsdf,mescheder2019occupancy,chen2019learning,sitzmann2020siren,atzmon2019sal,chabra2020deep,davies2020overfit,eslami2018neural,gropp2020implicit,michalkiewicz2019implicit,takikawa2021nglod,giebenhain2021airnets,boulch2022poco,mehta2021modulator} or use posed images to jointly optimize geometry and appearance \cite{chan2020pi,Chan2022eg3d,garbin2021fastnerf,hedman2021snerg,jiang2020sdfdiff,kellnhofer2021neural,lindell2020autoint,Liu2020neural,liu2019learning,liu2020dist,martinbrualla2020nerfw,mildenhall2020nerf,neff2021donerf,Niemeyer2020CVPR,oechsle2021unisurf,pumarola2020d,sitzmann2019srns,srinivasan2020nerv,yariv2020multiview,yu2021plenoctrees,zhang2020nerf}.
Neural fields represent scenes as continuous functions, allowing them to scale well with scene complexity compared to their discrete counterparts \cite{Lombardi:2019,sitzmann2019deepvoxels}.
Initial methods used a single, large multilayer perceptron (MLP) to represent entire scenes \cite{park2019deepsdf,mescheder2019occupancy,chen2019learning,mildenhall2020nerf,sitzmann2020siren}, but reconstruction with this approach can be computationally inefficient because training such a representation requires thousands of forward passes through the large model per scene.
Recent years have shown a trend towards locally conditioned representations, which either learn local functions~\cite{chabra2020deep,chen2021learning,jiang2020local,Reiser2021ICCV} or locally modulate a shared function with a hybrid explicit--implicit representation~\cite{Chan2022eg3d,peng2020convolutional,devries2021unconstrained,Liu2020neural,martel2021acorn,mehta2021modulator,giebenhain2021airnets,boulch2022poco,genova2019learning,genova2019deep}.
These methods use small MLPs, which are efficient during inference and significantly better at capturing local scene details.
We adopt the expressive hybrid triplane representation introduced by Chan et al.~\cite{Chan2022eg3d}.
Triplanes are efficient, scaling with the surface area rather than volume, and naturally integrate with expressive, fine-tuned 2D generator architectures.
We modify the triplane representation for compatibility with our denoising framework.

\paragraph{Generative synthesis in 2D and 3D.}
Some of the most popular generative models include GANs~\cite{goodfellow2014generative,karras2019style,karras2020stylegan2}, autoregressive models~\cite{Oord2016pixelrnn,Oord2017vqvae,Razavi2019vqvae2,Esser2020taming}, score matching models~\cite{Dickstein2015nonequilibrium_score,Song2019generative-gradients,Song2020score-improved-techniques}, and denoising diffusion probabilistic models (DDPMs)~\cite{Ho2020ddpm,Nichol2021improved-ddpm,Dhariwal2021ddpm-beat-gan,Vahdat2021score-latent}. DDPMs are arguably the SOTA approach for synthesizing high-quality and diverse 2D images~\cite{Dhariwal2021ddpm-beat-gan}. 
Moreover, GANs can be difficult to train and suffer from issues like mode collapse~\cite{thanhtung2020modecollapse} whereas diffusion models train stably and have been shown to better capture the full training distribution. 

In 3D, however, GANs still outperform alternative generative approaches~\cite{Liao2020unsupervised-3d-synthesis,Wu20163d-latent-gan,Nguyen-Phuoc2019hologan,Gadelha20173d-shape-induction,Niemeyer2021giraffe,Liu2020neural,chan2020pi, graf, Meng2021gnerf, Kosiorek2021nerf-vae, Rebain2022lolnerf,Chan2022eg3d,gu2021stylenerf,Zhou2021CIPS3D,Or-El2022style-sdf,zheng2022sdfstylegan}.
Some of the most successful 3D GANs use an expressive 2D generator backbone (e.g., StyleGAN2 \cite{karras2020stylegan2}) to synthesize triplane representations which are then decoded with a small, efficient MLP \cite{Chan2022eg3d}.
Because the decoder is small and must generalize across many local latents, these methods assign most of their expressiveness to the powerful backbone.
In addition, these methods treat the triplane as a multi-channel image, allowing the generator backbone to be used almost out of the box.

Current 3D diffusion models \cite{Luo2021ddpm-point-cloud,Zhou2021point-voxel-ddpm,Bautista2022gaudi,dupont2022functa,Yao2021dd-nerf} are still very limited. They either denoise a single latent or do not utilize neural fields at all, opting for a discrete point-cloud-based approach.
%
For example, concurrently developed single-latent approaches \cite{Bautista2022gaudi,dupont2022functa} generate a global latent for conditioning the neural field, relying on a 3D decoder to transform the scene representation from 1D to 3D without directly performing 3D diffusion. 
As a result, the diffusion model does not actually operate in 3D, losing this important inductive bias and generating blurry results.
Point-cloud-based approaches \cite{Luo2021ddpm-point-cloud,Zhou2021point-voxel-ddpm}, on the other hand, give the diffusion model explicit 3D control over the shape, but limit its resolution and scalability due to the coarse discrete representation.
While showing promise, both 1D-to-3D and point cloud diffusion approaches require specific architectures that cannot easily leverage recent advances in 2D diffusion models.

In our work, we propose to directly generate triplanes with out-of-the-box SOTA 2D diffusion models, granting the diffusion model near-complete control over the generated neural field.
Key to our approach is our treatment of well-fit triplanes in a shared latent space as ground truth data for training our diffusion model.
We show that the latent space of these triplanes is grounded spatially in local detail, giving the diffusion model a critical inductive bias for 3D generation.
Our approach gives rise to an expressive 3D diffusion model. 

\newcommand{\triplane}{\bold{f}}

\begin{figure*}[t]
    \centering
    \includegraphics[width=0.8\linewidth]{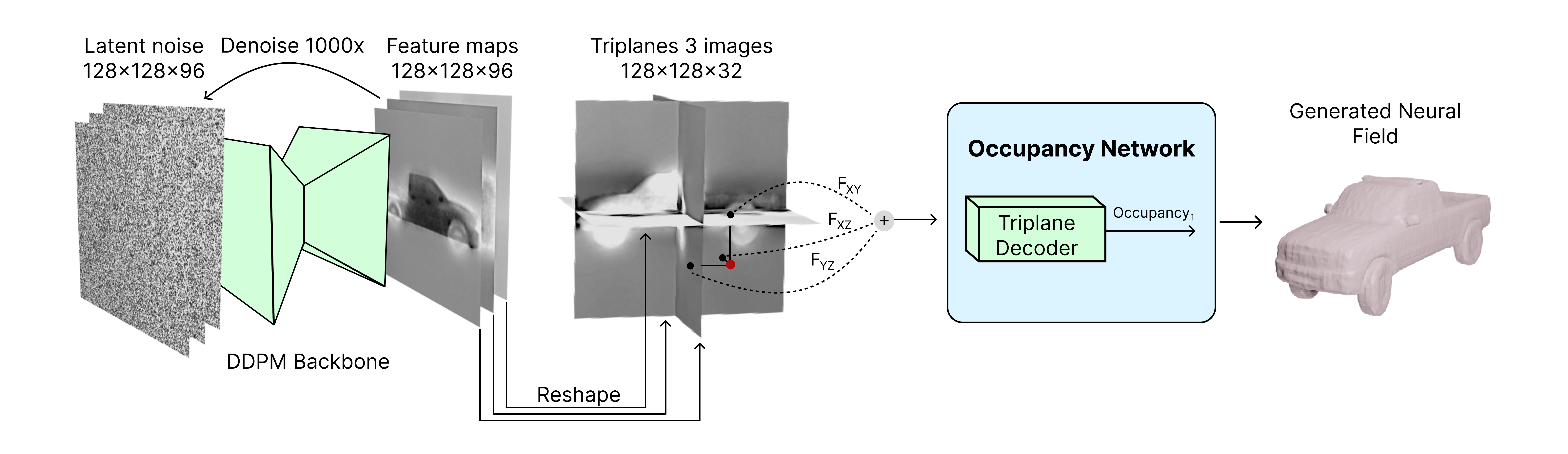}
    \caption{\textbf{Pipeline.} Sampling a 3D neural field from our model consists of two decoupled processes: 1) using a trained DDPM to iteratively denoise latent noise into feature maps and 2) using a locally conditioned Occupancy Network to decode the resulting triplane into the final neural field. This architecture allows the DDPM to generate samples with a 3D inductive bias while utilizing existing 2D DDPM backbones and a continuous output representation. 
    }
    \label{fig:diffusion-architecture}
\end{figure*}





\section{Triplane Diffusion Framework}

Here, we explain the architecture of our neural field diffusion (NFD) model for 3D shapes. In Section~\ref{sec:single_triplane}, we explain how we can represent the occupancy field of a single object using a triplane. In Section~\ref{sec:multiple_triplanes}, we describe how we can extend this framework to represent an entire dataset of 3D objects. In Section~\ref{sec:regularization}, we describe the regularization techniques that we found necessary to achieve optimal results. Finally, Sections~\ref{sec:training} and \ref{sec:sampling} illustrate training and sampling from our model. For an overview of the pipeline at inference, see Figure~\ref{fig:diffusion-architecture}.

\subsection{Representing a 3D Scene using a Triplane}
\label{sec:single_triplane}

Neural fields have been introduced as continuous and expressive 3D scene representations. In this context, a neural field $\textrm{\sc{nf}} : \mathbb{R}^{3} \rightarrow \mathbb{R}^{M}$ is a neural network--parameterized mapping function that takes as input a three-dimensional coordinate $\mathbf{x}$ and that outputs an $M$-dimensional vector representing the neural field. Neural fields have been demonstrated for occupancy fields~\cite{mescheder2019occupancy}, signed distance functions~\cite{park2019deepsdf}, radiance fields~\cite{mildenhall2020nerf}, among many other types of signals~\cite{sitzmann2020siren}. For the remainder of this work, we focus on 3D scene representations using occupancy fields such that the output of the neural field is a binary value, indicating whether a coordinate is inside or outside an object and $M=1$.

The triplane representation is a hybrid explicit--implicit network architecture for neural fields that is particularly efficient to evaluate~\cite{Chan2022eg3d}. This representation uses three 2D feature planes $\triplane_{xy},\triplane_{xz},\triplane_{yz} \in \mathbb{R}^{N \times N \times C}$ with a spatial resolution of $N \times N$ and $C$ feature channels each, and a multilayer perceptron (MLP) ``decoder'' tasked with interpreting features sampled from the planes. A 3D coordinate is queried by projecting it onto each of the axis-aligned planes (i.e., the $x\!-\!y$, $x\!-\!z$, and $y\!-\!z$ planes), querying and aggregating the respective features, and decoding the resulting feature using a lightweight $\textrm{\sc{mlp}}_\phi$ with parameters $\phi$. 
Similar to Chan et al.~\cite{Chan2022eg3d}, we found the sum to be an efficient feature aggregation function, resulting in the following formulation for the triplane architecture:
 \begin{equation}
 \textrm{\sc{nf}} \left( \mathbf{x} \right) = \textrm{\sc{mlp}}_\phi \left( \triplane_{xy} \left( \mathbf{x} \right) + \triplane_{yz}  \left( \mathbf{x} \right) + \triplane_{xz} \left( \mathbf{x} \right) \right).
 \end{equation}
The feature planes and $\textrm{\sc{mlp}}$ can be jointly optimized to represent the occupancy field of a shape.

\subsection{Representing a Class of Objects with Triplanes}
\label{sec:multiple_triplanes}

We aim to convert our dataset of shapes into a dataset of triplanes so that we can train a diffusion model on these learned feature planes. However, because the MLP and feature planes are typically jointly learned, we cannot simply train a triplane for each object of the dataset individually. If we did, the MLP's corresponding to each object in our dataset would fail to generalize to triplanes generated by our diffusion model. Therefore, instead of training triplanes for each object in isolation, we jointly optimize the feature planes for many objects simultaneously, along with a decoder that is \textit{shared} across all objects. This joint optimization results in a dataset of optimized feature planes and an MLP capable of interpreting any triplane from the dataset distribution. Thus, at inference, we can use this MLP to decode feature planes generated by our model.

In practice, during training, we are given a dataset of $I$ objects, and we preprocess the coordinates and ground-truth occupancy values of $J$ points per object. Typically, $J=10\text{M}$, where 5M points are sampled uniformly throughout the volume and 5M points are sampled near the object surface.
Our naive training objective is a simple $L2$ loss between predicted occupancy values $\textrm{\sc{nf}}^{(i)} ( \mathbf{x}^{(i)}_j )$ and ground-truth occupancy values $\textrm{\sc o}^{(i)}_j$
for each point, where $\mathbf{x}^{(i)}_j$ denotes the $j$\textsuperscript{th} point from the $i$\textsuperscript{th} scene:

\begin{equation} \label{eq:l_naive}
\mathcal{L}_\textrm{\sc{naive}} = 
\sum_i^I \sum_j^J
\left\| 
\textrm{\sc{nf}}^{(i)} \left( \mathbf{x}^{(i)}_j \right) - 
\textrm{\sc o}^{(i)}_j
\right\|_2
\end{equation}

During training, we optimize Equation~\ref{eq:l_naive} for a shared \textrm{\sc{MLP}} parameterized by $\phi$, as well as the feature planes corresponding to every object in our dataset:
\begin{equation}
    \left\{ \phi,\triplane_{xy}^{(i)},\triplane_{xz}^{(i)},\triplane_{yz}^{(i)} \right\} =\argmin_{ \left\{ \phi,\triplane_{xy}^{(i)},\triplane_{xz}^{(i)},\triplane_{yz}^{(i)} \right\}} \mathcal{L}_\textrm{\sc{naive}}
\end{equation}
%


\subsection{Regularizing Triplanes for Effective Generalization}
\label{sec:regularization}

Following the procedure outlined in the previous section, we can learn a dataset of triplane features and a shared triplane decoder; we can then train a diffusion model on these triplane features and sample novel shapes at inference. Unfortunately, the result of this naive training procedure is a generative model for triplanes that produces shapes with significant artifacts.

We find it necessary to regularize the triplane features during optimization to simplify the data manifold that the diffusion model must learn. Therefore, we include total variation (TV) regularization terms with weight $\lambda_1$ in the loss function to ensure that the feature planes of each training scene do not contain spurious high-frequency information. This strategy makes the distribution of triplane features more similar to the manifold of natural images (see supplement), which we found necessary to robustly train a diffusion model on them (see Sec.~\ref{sec:results}).

While the trained feature values are unbounded, our DDPM backbone requires training inputs with values in the range [-1,1]. We address this by normalizing the feature planes before training, but this process is sensitive to outliers. As a result, we include an L2 regularization term on the triplane features with weight $\lambda_2$ to discourage outlying values.

We also include an explicit density regularization (EDR) term. Due to our ground-truth occupancy data being concentrated on the surface of the shapes, there is often insufficient data to learn a smooth outside-of-shape volume. Our EDR term combats this issue by sampling a set of random points from the volume, offsetting the points by a random vector $\boldsymbol{\omega}$, feeding both sets through the MLP, and calculating the mean squared error. Notationally, this term can be represented as $\textrm{EDR} 
 \left( \textrm{\sc{nf}} \left( \mathbf{x} \right) , \boldsymbol{\omega} \right) = \|\textrm{\sc{nf}} \left( \mathbf{x} \right) - \textrm{\sc{nf}} \left( \mathbf{x}+\boldsymbol{\omega} \right)\|_2^2$. We find this term necessary to remove floating artifacts in the volume (see Sec.~\ref{sec:results})

Our training objective, with added regularization terms, is as follows:
\newcommand\numberthis{\addtocounter{equation}{1}\tag{\theequation}}
\begin{align*}
\mathcal{L}  = \sum_i^N \sum_j^M 
& \left\| \textrm{\sc{nf}}^{(i)} \left( \mathbf{x}^{(i)}_j \right) - \textrm{\sc o}^{(i)}_j \right\|_2  \\
 & + \lambda_1 \, \left(\textrm{TV} \left( \triplane^{(i)}_{xy} \right) + \textrm{TV} \left( \triplane^{(i)}_{xz} \right)  + \textrm{TV} \left( \triplane^{(i)}_{yz} \right)\right) \\
 & + \lambda_2 \, 
 \left(||\triplane^{(i)}_{xy}||_2 + ||\triplane^{(i)}_{yz}||_2 + ||\triplane^{(i)}_{xz}||_2\right) \\
 & + \textrm{EDR} 
 \left( \textrm{\sc{nf}} \left( \mathbf{x}^{(i)}_j \right) , \boldsymbol{\omega} \right)
 \numberthis \label{eq:l_full}
\end{align*}

\subsection{Training a Diffusion Model for Triplane Features}
\label{sec:training}







For unconditional generation, a diffusion model takes Gaussian noise as input and gradually denoises it in $T$ steps. In our framework, the diffusion model operates on triplane features $\triplane_{0 \ldots T} \in \mathbb{R}^{N \times N \times 3C}$ that stack the feature channels of all three triplane axes into a single image. In this notation, $\triplane_T \sim \mathcal{N} \left( \triplane_T; 0, \mathbf{I} \right)$ is the triplane feature image consisting of purely Gaussian noise, and $\triplane_0 \sim q \left( \triplane_0 \right)$ is a random sample drawn from the data distribution. The data distribution in our framework includes the pre-factored triplanes of the training set, normalized by the mean and variance of the entire dataset such that each channel has a zero mean and a standard deviation of 0.5.

The \emph{forward} or \emph{diffusion processes} is a Markov chain that gradually adds Gaussian noise to the triplane features, according to a variance schedule $\beta_1, \beta_2, \ldots, \beta_T$
\begin{equation}
    q \left( \triplane_t | \triplane_{t-1} \right) = \mathcal{N} \left( \triplane_t; \sqrt{1-\beta_t} \triplane_{t-1}, \beta_t \mathbf{I} \right).       
\end{equation}
This forward process can be directly sampled at step $t$ using the closed-form solution $q \left( \triplane_t | \triplane_0 \right) = \mathcal{N} \left( \triplane_t; \sqrt{\bar{\alpha}_t} \triplane_0, \left( 1 - \bar{\alpha}_t \right) \mathbf{I} \right)$, where $\bar{\alpha}_t = \prod_{s=1}^t \alpha_s$ with $\alpha_t = 1 - \beta_t$.


The goal of training a diffusion model is to learn the \emph{reverse process}. For this purpose, a function approximator $\boldsymbol{\epsilon}_\theta$ is needed that predicts the noise $\epsilon \sim \mathcal{N} \left( \mathbf{0}, \mathbf{I} \right)$ from its noisy input. Typically, this function approximator is implemented as a variant of a convolutional neural network defined by its parameters $\theta$. Following~\cite{Ho2020ddpm}, we train our triplane diffusion model by optimizing the simplified variant of the variational bound on negative log-likelihood:
\begin{equation}
    \mathcal{L}_\textrm{DDPM} \! = \! \mathbb{E}_{t,\triplane_0, \boldsymbol{\epsilon}} \left[ \left\| \boldsymbol{\epsilon} - \boldsymbol{\epsilon}_\theta \left(\sqrt{\bar{\alpha}_t} \triplane_0 + \sqrt{1-\bar{\alpha}_t} \boldsymbol{\epsilon}, t \right) \right\|^2 \right]\!\!,
\end{equation}
where $t$ is sampled uniformly between 1 and $T$.

\subsection{Sampling Novel 3D Shapes}
\label{sec:sampling}

The unconditional generation of shapes at inference is a two-stage process that involves sampling a triplane from the trained diffusion model and then querying the neural field.

Sampling a triplane from the diffusion model is identical to sampling an image from a diffusion model. Beginning with a random Gaussian noise $\triplane_T \sim \mathcal{N}(\mathbf{0}, \mathbf{I})$, we iteratively denoise the sample in $T$ steps as
\begin{equation}
    \triplane_{t-1} = \frac{1}{\sqrt{\alpha_t}} \left(\triplane_t - \frac{1 - \alpha_t}{\sqrt{1 - \bar{\alpha}}_t}\boldsymbol{\epsilon}_{\theta}(\triplane_t, t) \right) + \sigma_t \boldsymbol{\epsilon},
\end{equation}
where $\boldsymbol{\epsilon} \sim \mathcal{N}(\mathbf{0}, \mathbf{I})$ for all but the very last step (\ie, $t=1$), at which $\boldsymbol{\epsilon}=0$ and $\sigma_t^2=\beta_t$.

The result of the denoising process, $\triplane_0$, is a sample from the normalized triplane feature image distribution. Denormalizing it using the dataset normalization statistics and splitting the generated features into the axis aligned planes $\triplane_{xy},\triplane_{yz},\triplane_{xz}$ yields a set of triplane features which, when combined with the pre-trained $\textrm{\sc{mlp}}$, are used to query the neural field.


We use the marching cubes algorithm~\cite{lorensen1987marching} to extract meshes from the resulting neural fields. Note that our framework is largely agnostic to the diffusion backbone used; we choose to use ADM~\cite{Nichol2021improved-ddpm}, a 2D state-of-the-art diffusion model. 

Source code and pre-trained models will be made available. 

\section{Experiments}
\label{sec:results}

\begin{figure*}[h]
  \includegraphics[width=\textwidth]{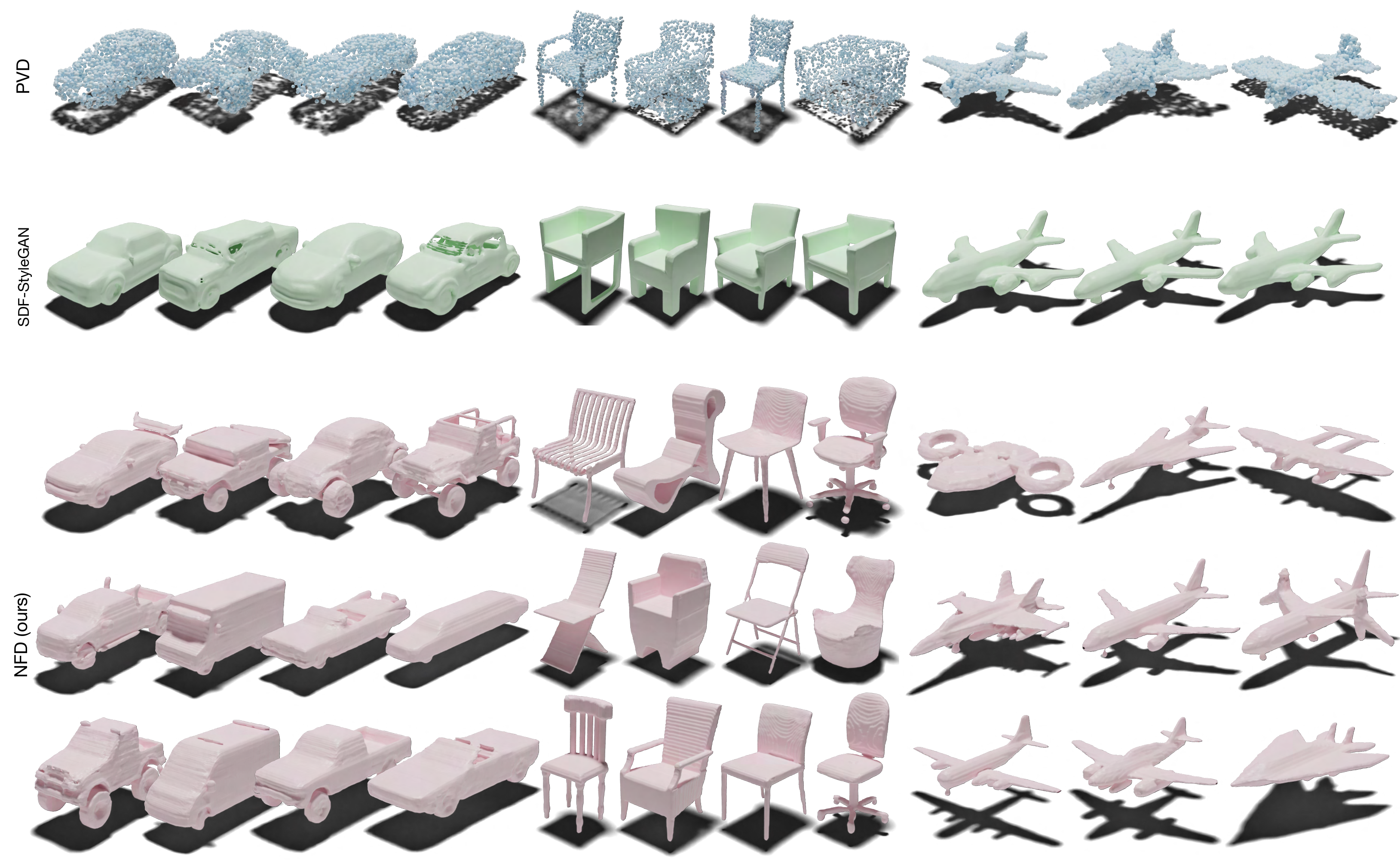}
  \caption{We compare 3D shapes generated by our model against generations of state-of-the-art baselines for ShapeNet \textit{Cars}, \textit{Chairs}, and \textit{Planes}. Our model synthesizes shapes with noticeably sharper details than the previous state-of-the-art, while also capturing the broad diversity in each category.}
  \label{fig:sample-results}
\end{figure*}

\begin{table}
  \centering
  \small
  \begin{tabular}{cccccc}
    \toprule
    Data & Method & FID $\downarrow$ & Precision $\uparrow$ & Recall $\uparrow$ \\
    \midrule
     & PVD$^*$ & 335.8 & 0.1  & 0.2 \\
    Cars & SDF-StyleGAN & 98.0 & 35.9 & 36.2\\
         & NFD (Ours)   & \textbf{83.6}   & \textbf{49.5} & \textbf{50.5}\\
    \midrule
     & PVD$^*$ & 305.8 & 0.2 & 1.7 \\
    Chairs & SDF-StyleGAN & 36.5 & 90.9 & 87.4\\
         & NFD (Ours)   & \textbf{26.4}   & \textbf{92.4} & \textbf{94.8}\\
    \midrule
     & PVD$^*$ & 244.4 &  2.7 &3.8\\
    Planes & SDF-StyleGAN & 65.8 & 64.5& 72.8 \\
         & NFD (Ours)   & \textbf{32.4}   & \textbf{70.5} & \textbf{81.1}\\
    \bottomrule
  \end{tabular}
  \caption{Render quality metrics on ShapeNet. We achieve state-of-the-art FID, which measures overall quality, as well as well as state-of-the-art precision and recall, which measure fidelity and diversity independently. Metrics calculated on shaded renderings of generated and ground-truth shapes.}
  \label{tab:main-results}
\end{table}

\paragraph{Datasets.} To compare NFD against existing 3D generative methods, we train our model on three object categories from the ShapeNet dataset individually. Consistent with previous work \cite{zheng2022sdfstylegan,Zhou2021point-voxel-ddpm}, we choose the categories: \textit{cars}, \textit{chairs} and \textit{airplanes}. Each mesh is normalized to lie within $[-1,1]^3$ and then passed through watertighting. The generation of ground truth triplanes then works as follows: we precompute the occupancies of 10M points per object, where 5M points are distributed uniformly at random in the volume, and 5M points are sampled within a 0.01 distance from the mesh surface. We then train an MLP jointly with as many triplanes as we can fit in the GPU memory of a single A6000 GPU. In our case, we initially train on the first 500 objects in the dataset. After this initial joint optimization, we freeze the shared MLP and use it to optimize the triplanes of the remaining objects in the dataset. All triplanes beyond the first 500 are optimized individually with the same shared MLP; thus, the training of these triplanes can be effectively parallelized.

\paragraph{Evaluation metrics.} As in~\cite{zheng2022sdfstylegan}, we choose to evaluate our model using an adapted version of Fr\'echet inception distance (FID) that utilizes rendered shading images of our generated meshes. Shading-image FID~\cite{zheng2022sdfstylegan} overcomes limitations of other mesh-based evaluation metrics such as the light-field-descriptor (LFD) \cite{Chen2003OnVS} by taking human perception into consideration. Zheng et al. \cite{zheng2022sdfstylegan} provide a detailed discussion of the various evaluation metrics for 3D generative models. Following the method~\cite{zheng2022sdfstylegan}, shading images of each shape are rendered from 20 distinct views; FID is then compared across each view and averaged to obtain a final score:
\begin{equation}
    \text{FID} =  \frac{1}{20}\left[\sum_{i=1}^{20} \|\mu_g^i - \mu_r^i\|^2 + \text{Tr}\left(\Upsigma_g^i + \Upsigma_r^i - 2(\Upsigma_r^i \Sigma_g^i)^{\frac{1}{2}}\right)\right],
\end{equation}
where $g$ and $r$ represent the generated and training datasets, while $\mu^i, \Upsigma^i$ represent the mean and covariance matrices for shading images rendered from the $i^\text{th}$ view, respectively.

Along with FID, we also report precision and recall scores using the method proposed by Sajjadi et al. \cite{Sajjadi2018AssessingGM}. While FID correlates well with perceived image quality, the one-dimensional nature of the metric prevents it from identifying different failure modes. Sajjadi et al. \cite{Sajjadi2018AssessingGM} aim to disentangle FID into separate metrics known as precision and recall, where the former correlates to the quality of the generated images and the latter represents the diversity of the generative model.

\paragraph{Baselines.}
We compare our method against state-of-the-art point-based and neural-field-based 3D generative models, namely PVD \cite{Zhou2021point-voxel-ddpm} and SDF-StyleGAN \cite{zheng2022sdfstylegan}. For evaluation, we use the pre-trained models for both methods on the three ShapeNet categories listed above. Note that PVD is inherently a point-based generative method and therefore does not output a triangle mesh needed for shading image rendering. To circumvent this, we choose to convert generated point clouds to triangle meshes using the ball-pivoting algorithm \cite{Bernardini1999TheBA}.

\paragraph{Results.} 
We provide qualitative results, comparing samples generated by our method to samples generated by baselines, in Figure~\ref{fig:sample-results}. Our method generates a diverse and finely detailed collection of objects. Objects produced by our method contain sharp edges and features that we would expect to be difficult to accurately reconstruct---note that delicate features, such as the suspension of cars, the slats in chairs, and armaments of planes, are faithfully generated. Perhaps more importantly, samples generated by our model are diverse—our model successfully synthesizes many different types of cars, chairs, and planes, including reproductions of several varieties that we would expect to be rare in the training dataset.

In comparison, while \emph{PVD} also produces a wide variety of shapes, it is limited by its nature to generating only coarse object shapes. Furthermore, because \emph{PVD} produces a fixed-size point cloud with only 2048 points, it cannot synthesize fine elements.

\begin{figure*}[t!]
    \centering
    \includegraphics[width=\textwidth]{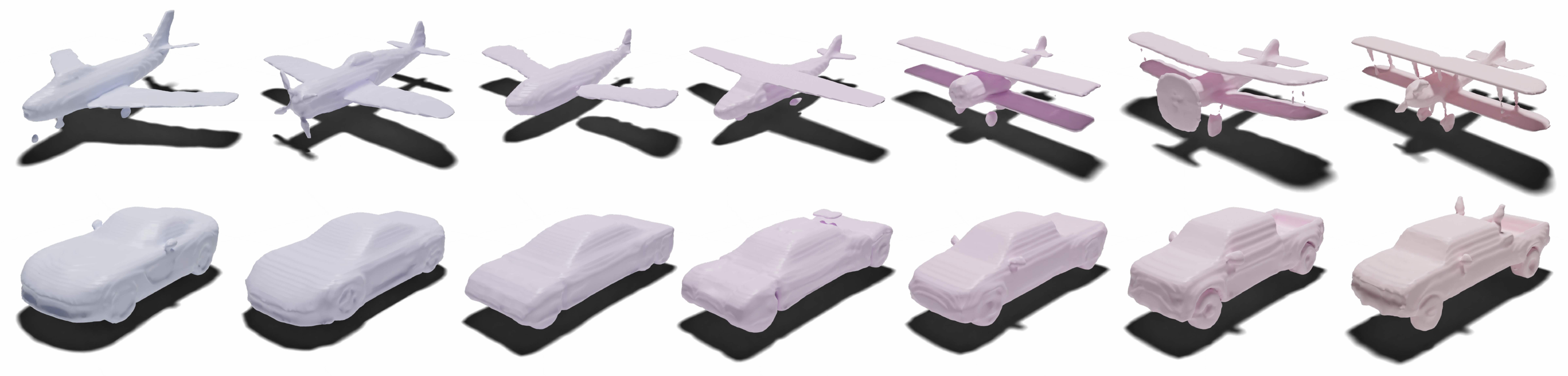}
    \caption{\textbf{Interpolation.} Our model learns a continuous latent space of triplanes. We can smoothly interpolate between two noise triplanes, resulting in semantically meaningful shape interpolation.}
    \label{fig:diffusion-interpolation}
\end{figure*}

\emph{SDF-StyleGAN} creates high-fidelity shapes, accurately reproducing many details, such as airplane engines and chair legs. However, our method is more capable of capturing very fine features. Note that while \emph{SDF-StyleGAN} smooths over the division between tire and wheel well when generating cars, our method faithfully portrays this gap. Similarly, our method synthesizes the tails and engines of airplanes, and the legs and planks of chairs, with noticeably better definition. Our method also apparently generates a greater diversity of objects than \emph{SDF-StyleGAN}. While \emph{SDF-StyleGAN} capably generates varieties of each ShapeNet class, our method reproduces the same classes with greater variation. This is expected, as a noted advantage of diffusion models over GANs is better mode coverage.

We provide quantitative results in Table~\ref{tab:main-results}. The metrics tell a similar story to the qualitative results. Quantitatively, NFD outperforms all baselines in FID, precision, and recall for each ShapeNet category. FID is a standard one-number metric for evaluating generative models, and our performance under this evaluation indicates the generally better quality of object renderings. Precision evaluates the renderings' fidelity, and recall evaluates their diversity. Outperforming baselines in both precision and recall suggest that our model produces higher fidelity of shapes and a more diverse distribution of shapes. This is consistent with the qualitative results in Figure~\ref{fig:sample-results}, where our method produced sharper and more complex objects while also covering more modes. 

\paragraph{Semantically meaningful interpolation.}

Figure~\ref{fig:diffusion-interpolation} shows latent space interpolation between pairs of generated neural fields. As shown in prior work~\cite{song2021denoising}, smooth interpolation in the latent space of diffusion models can be achieved by interpolation between noise tensors before they are iteratively denoised by the model. As in their method, we sample from our trained model using a deterministic DDIM, and we use spherical interpolation so that the intermediate latent noise retains the same distribution. Our method is capable of smooth latent space interpolation in the generated triplanes and their corresponding neural fields.


\subsection{Ablation Studies}


We validate the design of our framework by ablating components of our regularization strategies using the cars dataset.

\begin{figure}[t!]
  \includegraphics[width=0.5\textwidth]{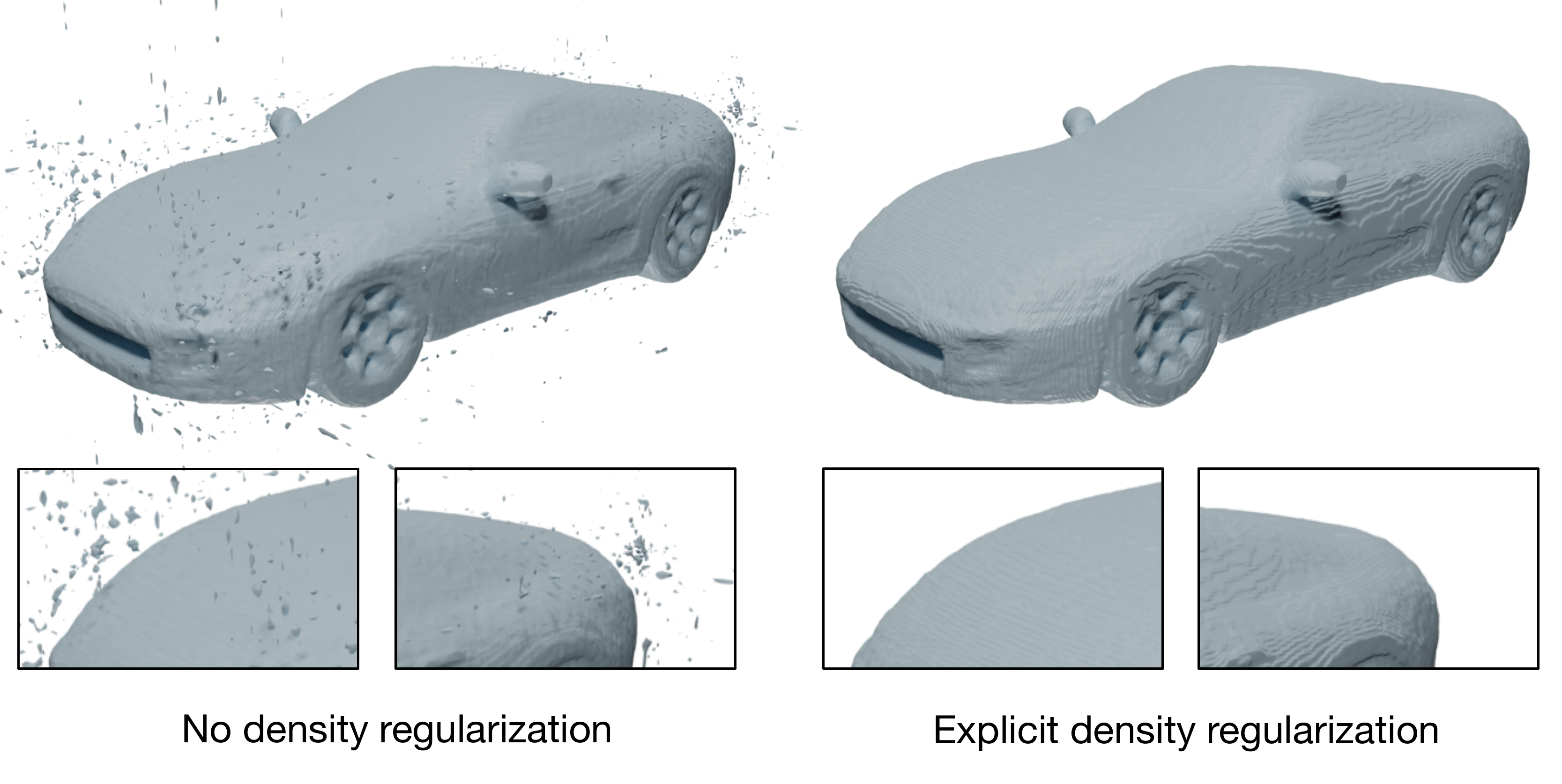}
  \caption{\textbf{Ablation over density regularization.} Clear artifacts are visible in the resulting occupancy field without explicit density regularization. In this example, we optimize a single triplane on a single shape.}
  \label{fig:edr}
\end{figure}

\paragraph{Explicit density regularization.} As discussed by Park et al. \cite{park2019deepsdf}, the precision of the ground truth decoded meshes is limited by the finite number of point samples guiding the training of the decision boundaries. Because we rely on a limited number of pre-computed coordinate--occupancy pairs to train our triplanes, it is easy to overfit to this limited training set. Even when optimizing a single triplane in isolation (i.e., \textit{without} learning a generative model), this overfitting manifests in ``floater'' artifacts in the optimized neural field. Figure \ref{fig:edr} shows an example where we fit a single triplane with and without density regularization. Without density regularization, the learned occupancy field contains significant artifacts; with density regularization, the learned occupancy field captures a clean object. 

\begin{figure}[t!]
  \includegraphics[width=0.5\textwidth]{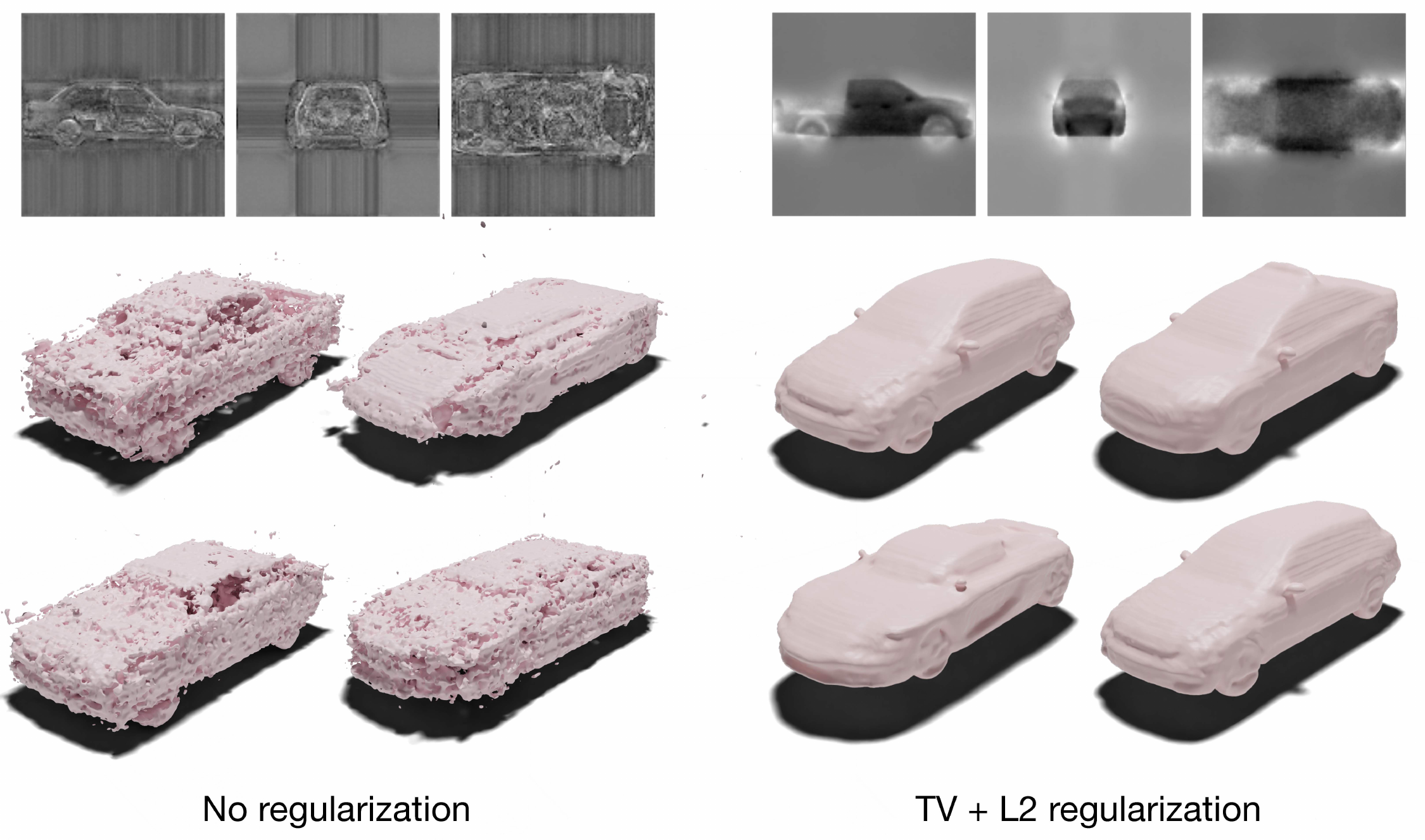}
  \caption{\textbf{Ablation over regularized triplanes.} A generative model trained on unregularized triplanes produces samples with significant artifacts. Effective regularization of triplane features enables training of a generative model that produces shapes without artifacts. Top left: triplane features learned only with Equation~\ref{eq:l_naive} contain many high frequency artifacts. Bottom left: a diffusion model trained on these unregularized triplanes fails to produce convincing samples. Top right: triplane features learned with Equation~\ref{eq:l_full} are noticeably smoother. Bottom right: A diffusion model trained on these regularized triplanes produces high-quality shapes.}
  \label{fig:triplane-reg}
\end{figure}


\paragraph{Triplane regularization.} Regularization of the triplanes is essential for training a well-behaved diffusion model. Figure \ref{fig:triplane-reg} compares generated samples produced by our entire framework, with and without regularization terms. If we train only with Equation~\ref{eq:l_naive}, i.e., without regularization terms, we can optimize a dataset of triplane features and train a diffusion model to generate samples. However, while the surfaces of the optimized shapes will appear real, the triplane features themselves will have many high-frequency artifacts, and these convoluted feature images are a difficult manifold for even a powerful diffusion model to learn. Consequently, generated triplane features produced by a trained diffusion model decode into shapes with significant artifacts. We note that these artifacts are present \textit{only} in generated samples; shapes directly factored from the ground-truth shapes are artifact-free, even without regularization.

Training with Equation~\ref{eq:l_full} introduces TV, L2, and density regularizing factors. Triplanes learned with these regularization terms are noticeably smoother, with frequency distributions that more closely align with those found in natural images (see supplement). As we would expect, a diffusion model more readily learns the manifold of regularized triplane features. Samples produced by a diffusion model trained on these regularized shapes decode into convincing and artifact-free shapes.


\section{Discussion}

In summary, we introduce a 3D-aware diffusion model that uses a 2D diffusion backbone to generate triplane feature maps, which are assembled into 3D neural fields. Our approach improves the quality and diversity of generated objects over existing 3D-aware generative models by a large margin.

\paragraph{Limitations.}
Similarly to other generative methods, training a diffusion model is slow and computationally demanding. Diffusion models, including ours, are also slow to evaluate, whereas GANs, for example, can be evaluated in real-time once trained. Luckily, our method will benefit from improvements to 2D diffusion models in this research area. Slow sampling at inference could be addressed by more efficient samplers~\cite{Karras2022ElucidatingTD} and potentially enable real-time synthesis.



\begin{figure}
    \centering
    \includegraphics[width=\linewidth]{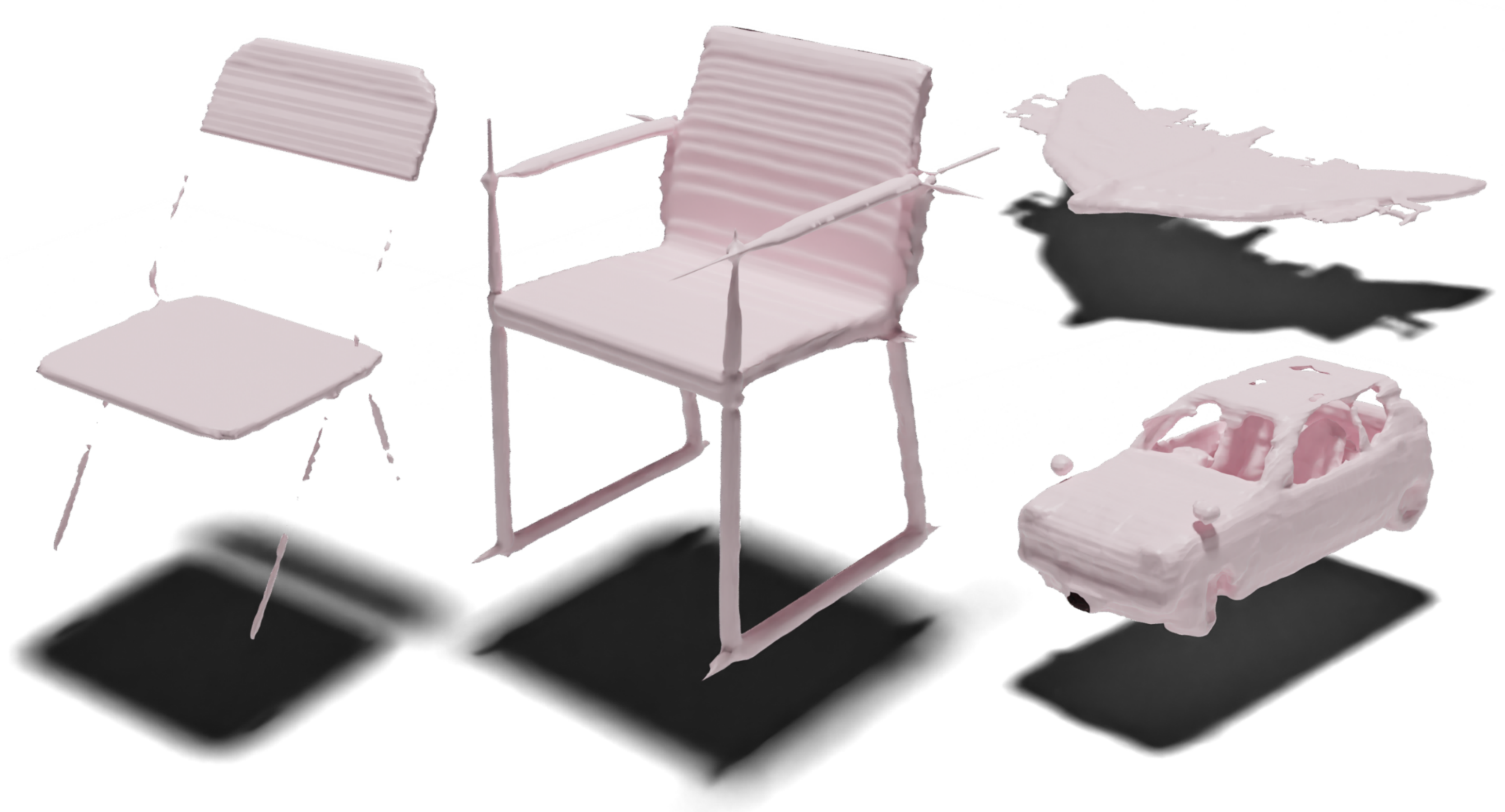}
    \caption{Failure cases.}
    \label{fig:my_label}
\end{figure}

\paragraph{Future Work.} 
We have demonstrated an effective way to generate occupancy fields, but in principle, our approach can be extended to generating any type of neural field that can be represented by a triplane. In particular, triplanes have already been shown to be excellent representations for radiance fields, so it seems natural to extend our diffusion approach to generating NeRFs.
While we demonstrate successful results for unconditional generation, conditioning our generative model on text, images, or other input would be an exciting avenue for future work.

\paragraph{Ethical Considerations.} 
Generative models, including ours, could be extended to generate DeepFakes. These pose a societal threat, and we do not condone using our work to generate fake images or videos of any person intending to spread misinformation or tarnish their reputation. 

\paragraph{Conclusion.}
3D-aware object synthesis has many exciting applications in vision and graphics. With our work, which is among the first to connect powerful 2D diffusion models and 3D object synthesis, we take a significant step towards utilizing emerging diffusion models for this goal.


\section*{Acknowledgements}

We thank Vincent Sitzmann for valuable discussions. This project was in part supported by Samsung, the Stanford Institute for Human-Centered AI (HAI), the Stanford Center for Integrated Facility Engineering (CIFE), NSF RI \#2211258, Autodesk, and a PECASE from the ARO.


{\small
\bibliographystyle{ieee_fullname}
\bibliography{main.bib}
}

\end{document}


\title{3D Neural Field Generation using Triplane Diffusion}

\maketitle

\tableofcontents

\section{Implementation Details}

\subsection{Learning a Dataset of Triplane Features}

\paragraph{Data.}
We train our model on 3 separate categories from the ShapeNet V1 dataset: \emph{Cars}, which contains 7496 objects, \emph{Chairs}, which contains 4971 objects, and \emph{Planes}, which contains 4045 objects. We train a separate model for each class of objects.
\paragraph{Watertighting.}
As a preprocessing step, we convert meshes from the ShapeNet dataset into watertight meshes. We perform watertighting with the implementation and settings from Mescheder et al. \cite{Mescheder2019Occupancy}. We render depth images from 20 views from a dodecahedron, which gives equally spaced views, and use the marching cubes algorithm~\cite{lorensen1987marching} to extract a watertight mesh.

\paragraph{Computing ground truth occupancy.}
We follow the implementation of \cite{Mescheder2019Occupancy} for computing occupancy values for arbitrary 3D coordinates. For any point in 3D space, we compute the occupancy value of the point by casting a ray along the $z$-axis and counting the number of intersections with the watertight mesh—an odd number of intersections means the point is inside the watertight shape.
When computing our dataset, we draw half of our query points uniformly at random from the volume, while the rest are importance sampled near the surface of the watertight mesh.

\paragraph{Triplane Features}
We used triplane features of dimension $128\times128\times32\times3$. While higher triplane resolutions guarantee lower degradation of decoded ground truth meshes, the increased dimensionality also places a burden on time and memory constraints. We initialize the triplane features to values drawn from a normal distribution with standard deviation 0.1.

\paragraph{Shared MLP.}
Our MLP is designed to be lightweight to enable quick training and inference. Our MLP is composed of a Fourier feature mapping layer \cite{Tancik2020FourierFL} with a scale factor of 1, followed by 3 fully connected layers of dimension 128, each with ReLU activation functions.

\paragraph{Training.}
As discussed in the main manuscript, we train our triplanes and MLP in two stages: first jointly on a subset of data, then independently on each object in the dataset, with a frozen MLP. During the first stage, we train on 500 randomly selected shapes with a batch size of 1 object per iteration and 500k occupancy values points per object. We train this first stage for 200 epochs with a learning rate of 1e-3. Training was conducted on a single RTX 2080ti, and took approximately 1 day to complete. The shared MLP is then frozen and used to train triplane features for every object object in the dataset. During this second stage, we train triplane features for each object individually. We use a batch size of 200k occupancy values per object and train for 30 epochs with a learning rate of 1e-3. This stage takes 10 minutes to train on an RTX 2080ti per shape, but can be parallelized across an arbitrary number of GPUs. The resulting triplane features are used as pseudo-ground truth images for training the diffusion model.

\subsection{Training a Diffusion Model for Triplane Features}

We base our implementation on the official code-base of \cite{Dhariwal2021ddpm-beat-gan}, available at \url{https://github.com/openai/guided-diffusion}. Unless otherwise stated, DDPM hyperparameters are identical to the class-specific LSUN model in \cite{Dhariwal2021ddpm-beat-gan}.

\paragraph{Diffusion Model Training.}
We train all models with a batch size of 128 across 8 A6000 GPUs. For cars, we used a learning rate of 1e-4 while for chairs and planes, a lower learning rate of 3e-5 helped prevent instability during training. We trained cars, chairs, and planes for 400k, 200k, and 200k steps respectively. Cars took around 6 days to train while chairs and planes each took approximately 3 days. The cars model was pretrained on a subset of the cars data for 160k steps before training on the full dataset for the remaining 240k iterations. 

\paragraph{Normalization.}
The learned triplane feature images, with which we train our diffusion model, are regularized (see Sec. \ref{triplane-reg}) but still theoretically unbounded, and we find outliers to skew the distribution. We apply normalizization to ensure the values of the triplane feature images to be within a fixed range. We normalize the feature channels to zero-mean and clip each channel to be within $S=16$ standard deviations of the mean. We then scale each channel to be within the range $[-1, 1]$.

\paragraph{Sampling at inference.}
When generating shapes, we default to using a DDPM with 1000 iterations. Generating a set of triplane features for a single example takes roughly 20 seconds on a single A6000 GPU, but the number of iterations can be decreased to 250 for faster generation and a small (judged visually) reduction in fidelity. Decoding the resulting occupancy field and extracting a mesh at a resolution of $128^3$ takes about 5 seconds per mesh, including both MLP evaluation and marching cubes.

\paragraph{Interpolation.}
We used DDIM \cite{song2021denoising} to sample shapes for interpolation. We noticed visually worse-quality meshes in the DDIM setting compared to the DDPM setting. Cars, chairs, and planes were sampled with 25, 250, and 25 steps respectively, though we noticed only small differences when the number of steps was changed.

\newpage

\section{Triplane Regularization}\label{triplane-reg}
\begin{figure}[h]
    \centering
    \includegraphics[width = 0.6\linewidth]{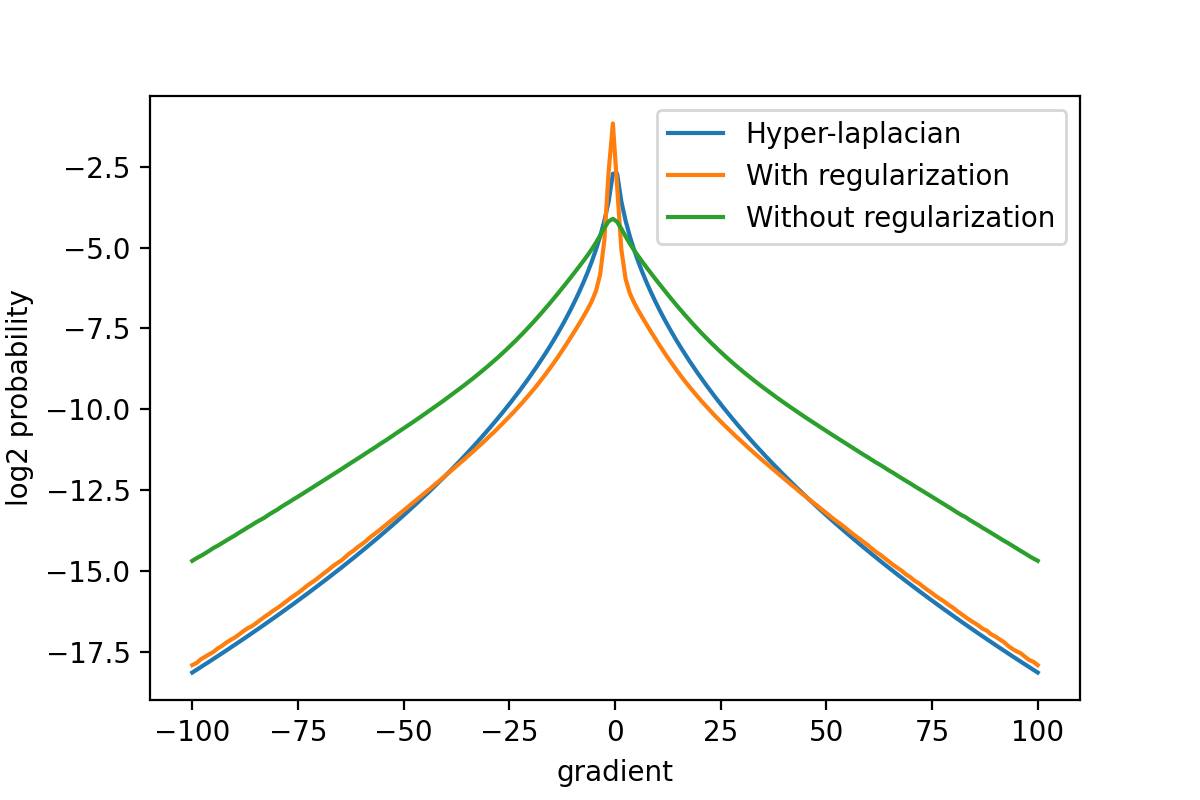}
    \caption{Distribution of image gradients for natural images and triplane features with and without regularization. After regularization, image gradients of ground truth triplane features closely resembles gradients found in natural images. Natural image gradients modelled by a hyper-Laplacian with $\alpha = 0.5$ per Krishnan et al. \cite{Krishnan2009FastID}.}
    \label{fig:gradients}
\end{figure}
State-of-the-art diffusion models have empirically performed well when trained on natural images. However, without proper regularization, ground truth triplanes trained using an autodecoder result in high frequency artifacts as shown in Figure 7. We apply TV regularization as illustrated in Equation 4, resulting in smoother triplane features that are more similar to the manifold of natural images.

Krishnan et al. \cite{Krishnan2009FastID} found that gradients of natural images are closely modelled by a hyper-Laplacian with $0.5 \leq \alpha \leq 0.8$. Supplementary Figure \ref{fig:gradients} shows the distribution of gradients of natural images modelled by a hyper-Laplacian with $\alpha = 0.5$ and gradients of trained triplane features with and without TV regularization. Gradients of triplanes trained with regularization closely resemble gradients found in natural images.

\newpage

\section{Generated Samples}
\begin{figure}[h]
    \centering
    \includegraphics[height=20cm]{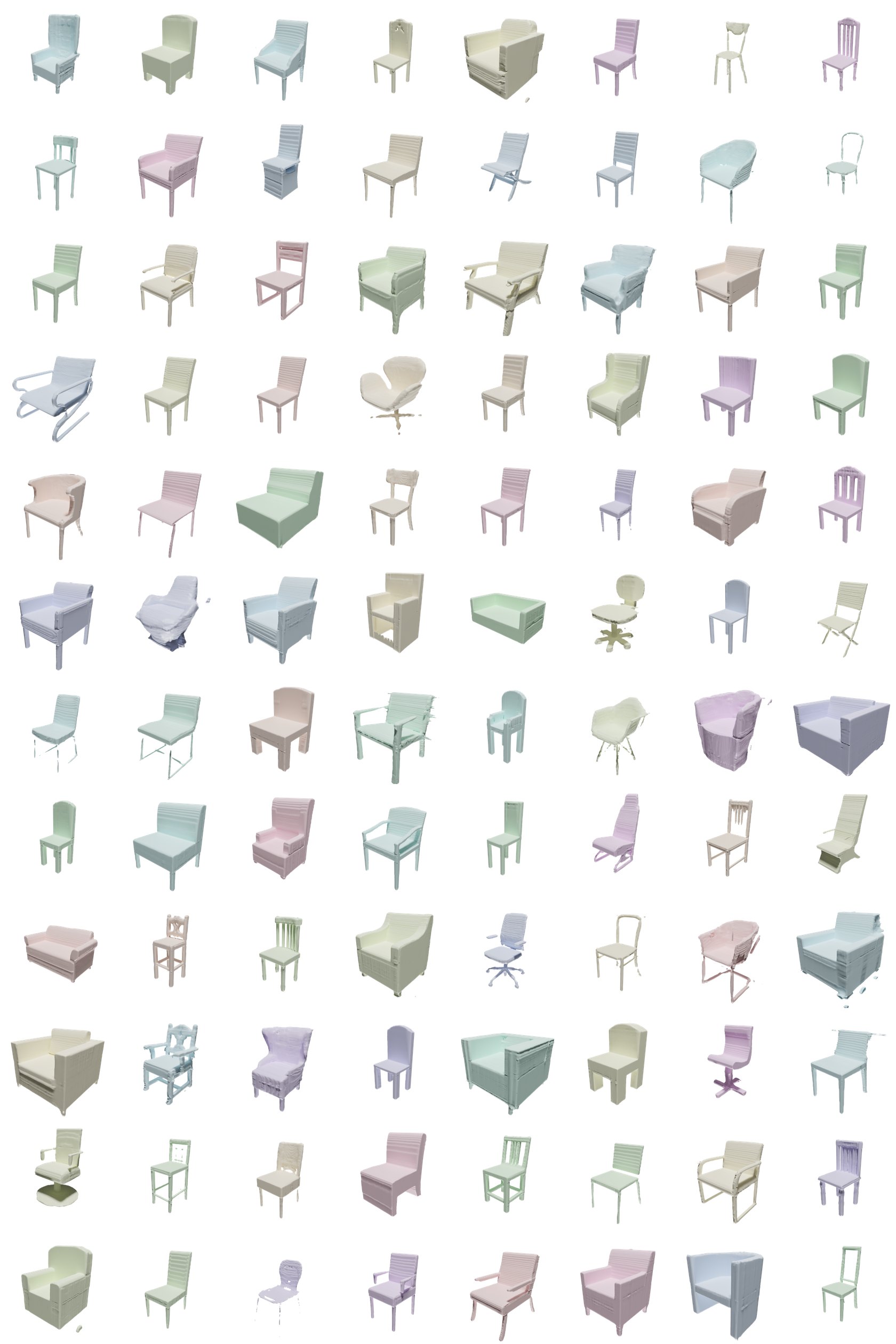}
    \caption{Set of 96 uncurated samples generated from our model trained on the \textit{chairs} category of ShapeNet.}
    \label{fig:chair_samples}
\end{figure}

\begin{figure}
    \centering
    \includegraphics[width=\linewidth]{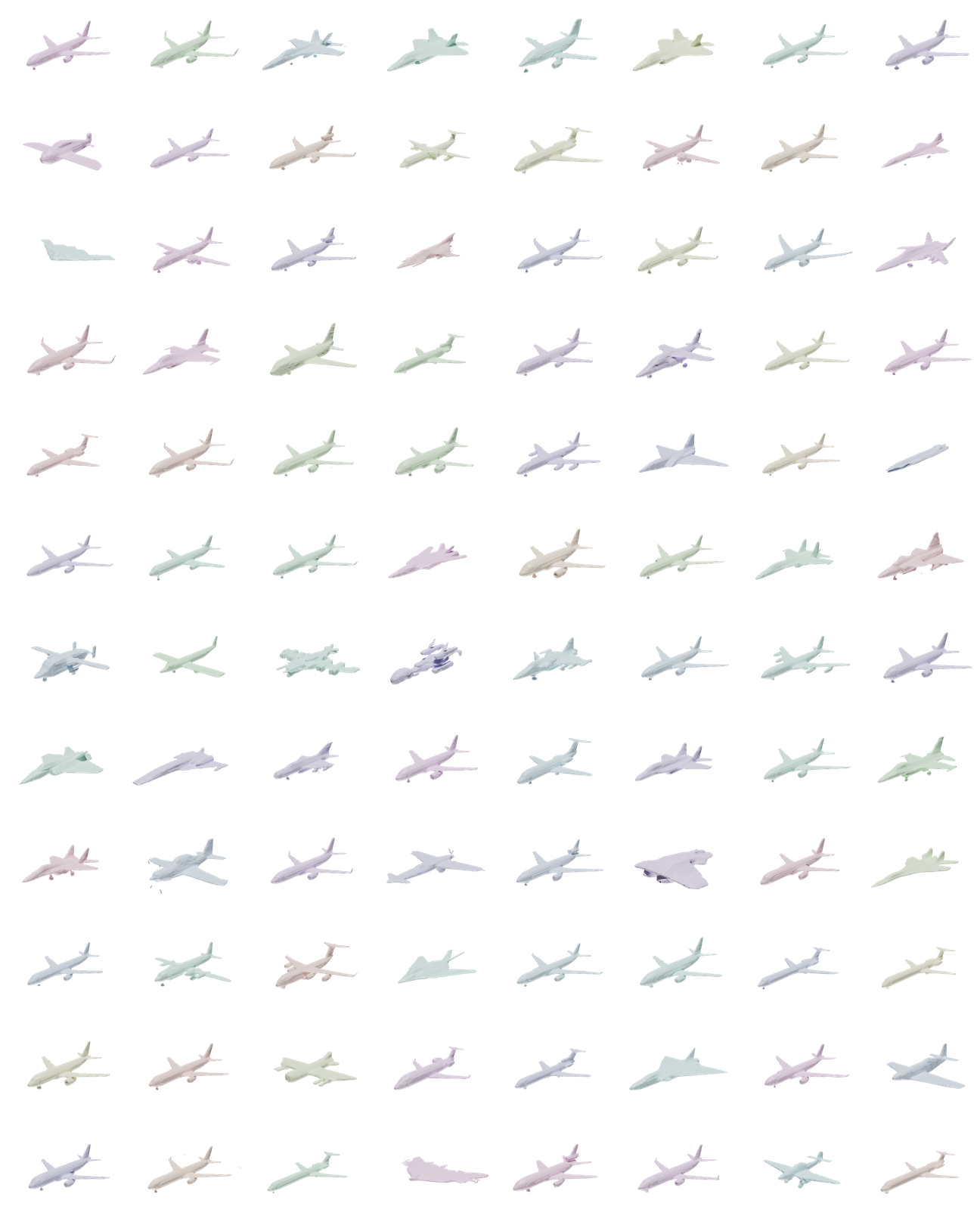}
    \caption{Set of 96 uncurated samples generated from our model trained on the \textit{planes} category of ShapeNet.}
    \label{fig:plane_samples}
\end{figure}

\begin{figure}
    \centering
    \includegraphics[width=\linewidth]{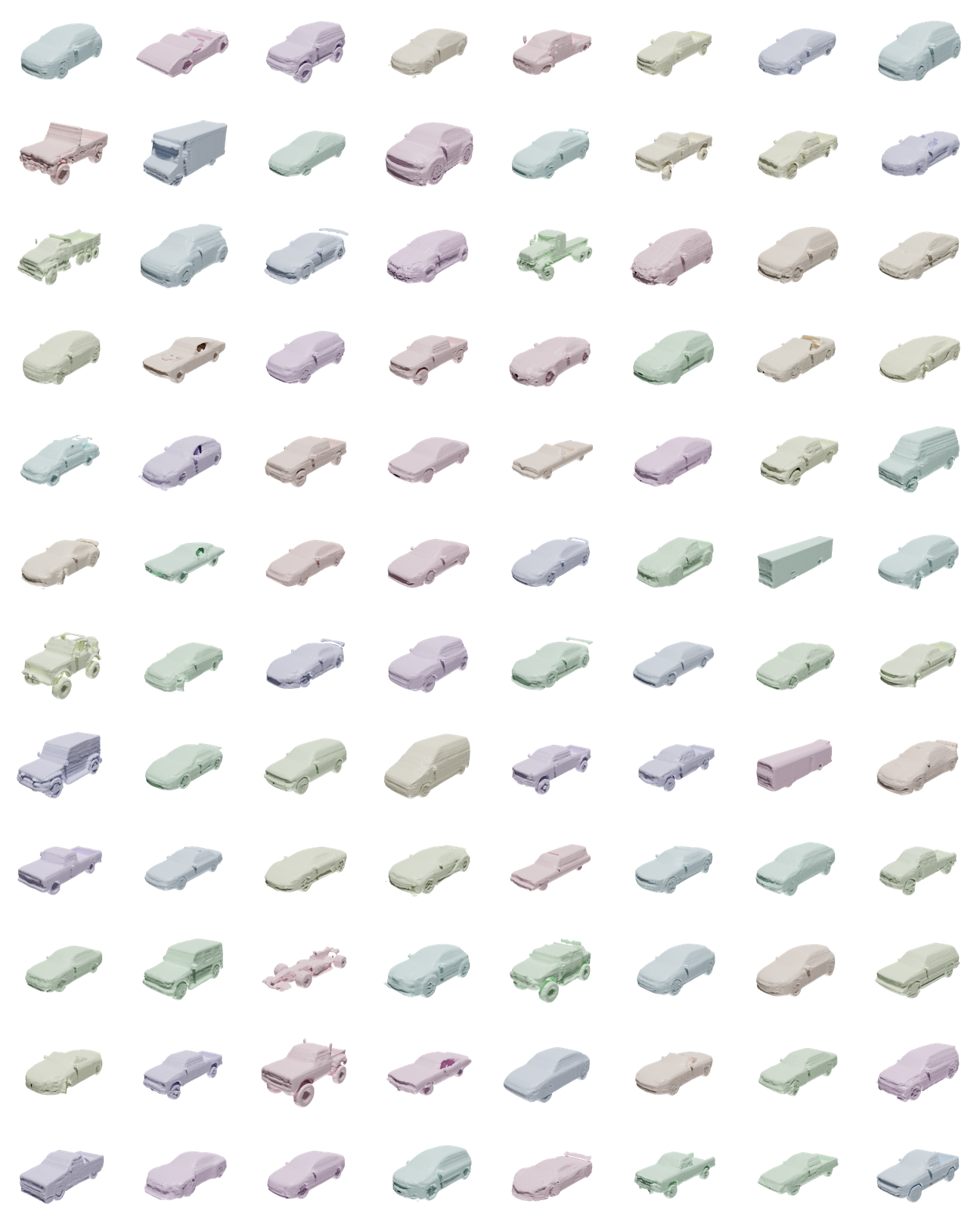}
    \caption{Set of 96 uncurated samples generated from our model trained on the \textit{cars} category of ShapeNet.}
    \label{fig:car_samples}
\end{figure}

\section{Comparison to Implicit-Grid 
\cite{Ibing20213DSG} Baseline}

\begin{figure}[h]
    \centering
    \includegraphics[width = \linewidth]{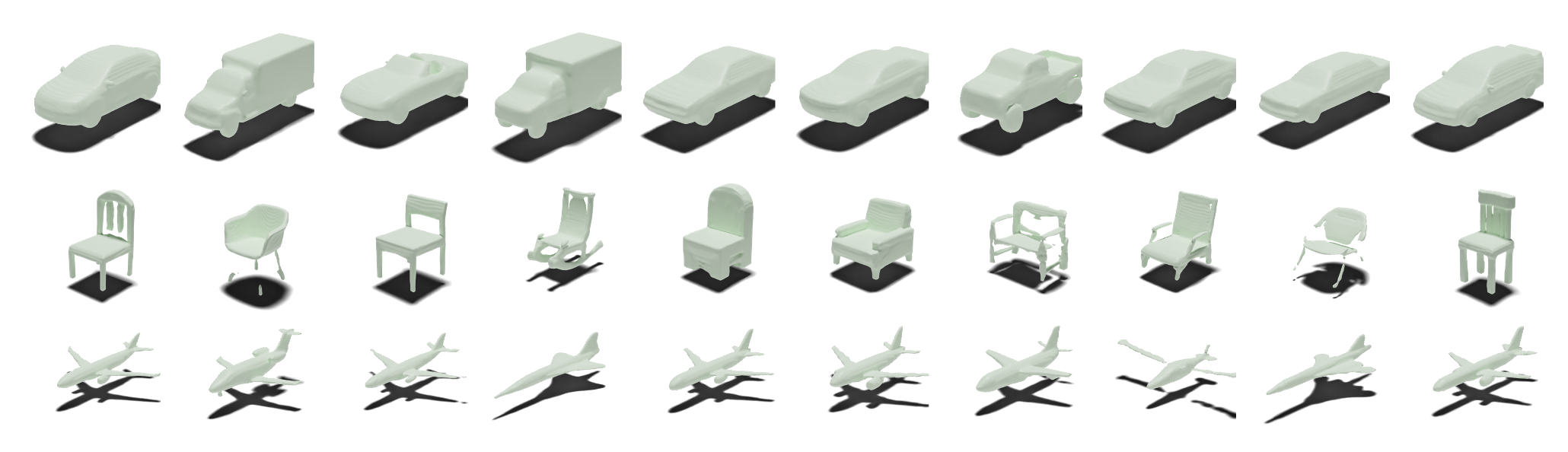}
    \caption{Generated shapes using Implicit-Grid baseline method \cite{Ibing20213DSG}.}
    \label{fig:imp_samples}
\end{figure}

\begin{table}[h]
  \centering
  \small
  \begin{tabular}{cccccc}
    \toprule
    Data & Method & FID $\downarrow$ & Precision $\uparrow$ & Recall $\uparrow$ \\
    \midrule
     & PVD$^*$ & 335.8 & 0.1  & 0.2 \\
    & Implicit-Grid & 209.3 & 25.9 & 21.5\\
    Cars & SDF-StyleGAN & 98.0 & 35.9 & 36.2\\
         & NFD (Ours)   & \textbf{83.6}   & \textbf{49.5} & \textbf{50.5}\\
    \midrule
     & PVD$^*$ & 305.8 & 0.2 & 1.7 \\
     & Implicit-Grid & 119.5 & 74.8 & 77.2\\
    Chairs & SDF-StyleGAN & 36.5 & 90.9 & 87.4\\
         & NFD (Ours)   & \textbf{26.4}   & \textbf{92.4} & \textbf{94.8}\\
    \midrule
     & PVD$^*$ & 244.4 &  2.7 &3.8\\
     & Implicit-Grid & 145.4 & 67.1 & 66.2\\
    Planes & SDF-StyleGAN & 65.8 & 64.5& 72.8 \\
         & NFD (Ours)   & \textbf{32.4}   & \textbf{70.5} & \textbf{81.1}\\
    \bottomrule
  \end{tabular}
  \caption{Comparison of evaluation metrics with baseline methods. Our method outperforms all baselines in FID, precision and recall, ullustrating that our method generates high quality and diverse 3D shapes.}
  \label{tab:main-results}
\end{table}

{\small
\bibliographystyle{ieee_fullname}
\bibliography{nfd_supplement}
}